\documentclass[sigconf]{acmart}

\AtBeginDocument{%
  \providecommand\BibTeX{{%
    \normalfont B\kern-0.5em{\scshape i\kern-0.25em b}\kern-0.8em\TeX}}}

\copyrightyear{2024}
\acmYear{2024}
\setcopyright{rightsretained}
\acmConference[CUI '24]{ACM Conversational User Interfaces 2024}{July 8--10, 2024}{Luxembourg, Luxembourg}
\acmBooktitle{ACM Conversational User Interfaces 2024 (CUI '24), July 8--10, 2024, Luxembourg, Luxembourg}
\acmDOI{10.1145/3640794.3665577}
\acmISBN{979-8-4007-0511-3/24/07}

\begin{document}

\title{Automatic Generation of Conversational Interfaces for Tabular Data Analysis}

\author{Marcos Gomez-Vazquez}
\email{marcos.gomez@list.lu}
\orcid{0000-0001-7176-0793}
\affiliation{%
  \institution{Luxembourg Institute of Science and Technology}
  \city{Esch-sur-Alzette}
  \country{Luxembourg}
  \postcode{4362}
}

\author{Jordi Cabot}
\email{jordi.cabot@list.lu}
\orcid{0000-0003-2418-2489}
\affiliation{%
  \institution{Luxembourg Institute of Science and Technology}
  \city{Esch-sur-Alzette}
  \country{Luxembourg}
  \postcode{4362}}
\affiliation{%
  \institution{University of Luxembourg}
  \city{Esch-sur-Alzette}
  \country{Luxembourg}
  \postcode{4362}}

\author{Robert Clarisó}
\email{rclariso@uoc.edu}
\orcid{0000-0001-9639-0186}
\affiliation{%
  \institution{Universitat Oberta de Catalunya}
  \city{Barcelona}
  \country{Spain}
  \postcode{43017-6221}
}

\renewcommand{\shortauthors}{Gomez-Vazquez, et al.}

\begin{abstract}
Tabular data is the most common format to publish and exchange structured data online. A clear example is the growing number of open data portals published by public administrations. However, exploitation of these data sources is currently limited to technical people able to programmatically manipulate and digest such data. As an alternative, we propose the use of chatbots to offer a conversational interface to facilitate the exploration of tabular data sources, including support for data analytics questions that are responded via charts rendered by the chatbot. Moreover, our chatbots are automatically generated from the data source itself thanks to the instantiation of a configurable collection of conversation patterns matched to the chatbot intents and entities.
\end{abstract}



\received{18 April 2024}
\received[revised]{23 May 2024}

\maketitle

\section{Introduction}

Tabular data, consisting of samples (rows) and features (columns), is a prevalent data type in digital technology, increasingly used in \emph{open data} published by public administrations\footnote{Just the EU portal \url{https://data.europa.eu/} registers over 1.5M datasets}. Despite its wide use, there's a significant lack of tools that allow non-technical users to easily explore this data, which limits the public advantage from open data initiatives. Conversational User Interfaces (CUIs) such as chatbots and voicebots 
could improve the accessibility of tabular data \cite{accessibility}. Until now, chatbots for tabular data are either manually created (an option that is not scalable) or completely relying on general purpose LLMs (with limited capacity, especially for larger datasets, and with a risk of generating wrong answers, see Section \ref{section:related-work}). 

This paper introduces a scalable, no-code tool that automatically creates chatbots for tabular data based on a schema inferred from the data itself. Such schema can be optionally enhanced by the user, or automatically with the help of a Large Language Models (LLMs) if needed. Our generated chatbots can handle a broad range of user intents and incorporate LLMs for generating responses through English-to-SQL translations when the intent corresponding to the user question is unrecognized. The entire setup is managed on our DataBot platform\footnote{\url{https://github.com/BESSER-PEARL/databot}}, which supports data import, chatbot management, and interactions via text or voice, providing outputs in tables or graph formats. This innovative approach not only simplifies the exploration and usage of tabular data for users without technical expertise but also offers organizations a convenient, effective method to enhance the value and utility of their data assets.



\section{Chatbot Architecture}
\label{section:chatbot-outline}

\begin{figure}[ht!]
    \centering
    \includegraphics[width=\linewidth]{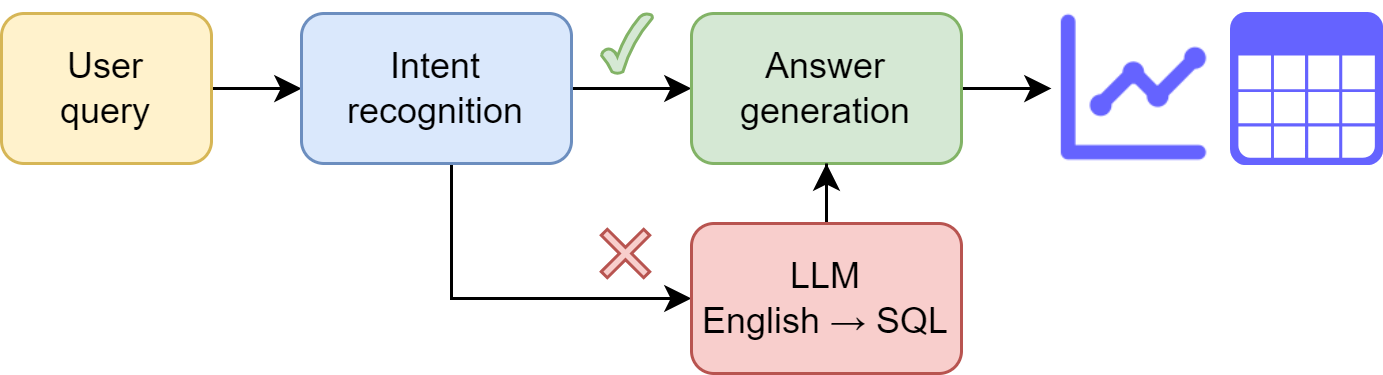}
    \caption{Diagram of the architecture of the generated chatbots.}
    \Description{Diagram of the architecture of the generated chatbots.}
    \label{fig:bot-architecture}
\end{figure}

The architecture we propose for our generated chatbots is depicted in Figure \ref{fig:bot-architecture}. At the core of the bot, there is an intent matching process aimed at identifying the user questions and their parameters. If the bot is able to match the intent and recognize all the mandatory parameters for that specific intent, it will transition to the appropriate state in charge of generating the answer for that particular question. If not, a fallback mechanism is triggered and an English-to-SQL translation is done by a LLM to obtain the best possible tabular answer. Note also that the bot is prepared to be multilingual with minimal work. 


\subsection{Execution Path for Recognized Intents}
\label{section:recognized-intent-path}
Once the chatbot knows the user's intent, it starts the process of generating the answer. This overall process is the same in all intents, though its implementation varies on each intent.

The first step is to analyze the intent parameters (entities). If the extracted parameters are OK (there are no missing parameters or wrong value types) the bot moves to the next step of the workflow. When the bot finds confusing or missing parameters, it triggers an interaction with the user to clarify the doubts. If the problem persists, it moves to the fallback state.

When all parameters are OK, the bot proceeds with the answer generation. At this point, the bot knows exactly what kind of question the user is asking (based on the recognized intent) and the variables involved in the question (the recognized parameters). The bot can generate 2 kinds of answers: tables or charts (depending on the type of information the user asks for, either implicitly or explicitly). In both cases, the bot behaves in a deterministic way, building the right SQL queries based on the detected information and reporting it in the desired output format (\emph{e.g.}, generating a pie chart and displaying it in the interactive GUI of the bot).

\subsection{LLM Fallback Path}

Despite the bot's best efforts, sometimes it may fail to understand the user's question, either because it is not one of the questions the tool foresaw in the generation or because it is too far from the training sentences. When this happens, the bot cannot give an exact answer on its own. At this point, it could just tell the user that it was unable to understand the question, but we try to be more useful and add to the bot a powerful, optional and configurable fallback mechanism.

The fallback relies on a LLM to automatically translate the user's query to an equivalent SQL statement. 
This approach does not always provide a perfect translation, and therefore, it may generate a wrong answer but it is worth trying as our experiments suggest that users prefer an approximate result (even if potentially wrong) than a plain ``sorry'' message. Note that the bot always warns the user when answering via this fallback strategy and provides the SQL suggested by the LLM to the user for explainability purposes. The quality of the answers will depend on the selected LLM. 
As of today, we obtain the best results with GPT-4 \cite{openai2023gpt4}. When triggering the fallback mechanism, the bot will: 1 - Send a prompt to the chosen LLM with instructions on the task it must perform (translation of a query in English into an equivalent SQL query) and the data schema in JSON format as additional context, 2 - Receive the LLM-generated SQL query and run it on the bot's dataset, and 3 - Display the tabular answer together with the SQL query used to obtain it.

\section{Automated Chatbot Generation Process}
\label{section:automated-chatbot-generation}

\begin{figure}[ht!]
    \centering
    \includegraphics[width=\linewidth]{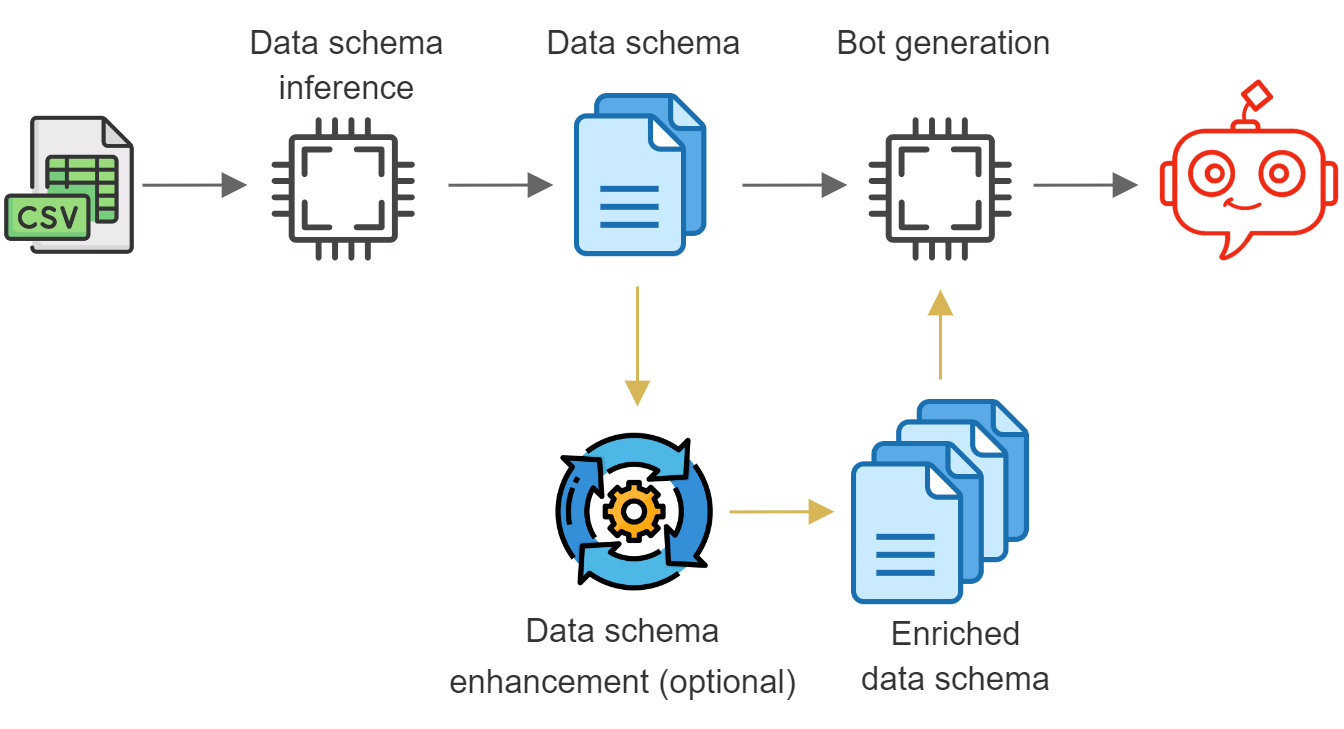}
    \caption{The chatbot generation process.}
    \Description{The chatbot generation process.}
    \label{fig:bot-generation-process}
\end{figure}

Figure~\ref{fig:bot-generation-process} shows the workflow our tool follows to generate the bots from an initial tabular data source, depicted as a CSV file in the Figure. The generated bots will follow the architecture explained in the previous section. The process is fully automatic, although the data owner can optionally participate in the data schema enrichment step to generate more powerful bots. This enrichment can also be automated by using LLMs. The next subsections describe in more detail each step.

\subsection{Data Schema Inference}
\label{section:feature-inference}

To automatically create a bot, we only need one ingredient: a tabular dataset. The dataset must follow a 2D structure, being composed by columns (attributes) and rows (records). Therefore, valid formats for this approach include CSV or XLSX, although other formats such as JSON or XML can be supported as long as they follow a tabular-like structure (\emph{e.g.}, nested attributes are not supported). The size of the dataset may impact the bot performance. The dataset queries will be done with Python's Pandas library, so the platform scalability strongly relies on Pandas' scalability capabilities (\emph{e.g.}, to execute several bots on top of  gigabytes of data, a certain level of computational power and memory will be necessary to run them)

From the structure of the dataset we will gather the list of columns/fields (with their names). From the analysis of the dataset content, we will infer the data type of each field (numeric, textual, date-time,...) and its diversity (number of different values present in that specific column). Based on a predefined (but configurable) diversity threshold, we automatically classify as \textit{categorical} those fields under the threshold. Categorical fields are implemented as an additional bot entity so that users can directly refer to their values in any question. All this information 
conforms the metadata the bot will be trained on.

At this point, the chatbot can be already generated. Thanks to the schema inferred from the data source, the chatbot will be able to recognise certain kinds of questions relying on the knowledge extracted from the data (\emph{e.g.}, ``Which is the maximum value in X?'', where X would correspond to one of the detected numeric columns'). Think of fields and rows as input and output parameters of the user's questions the bot must be able to answer, \emph{e.g.}, users can ask for the value in field \emph{X} of rows satisfying a certain condition in field \emph{Y}.

\subsection{Data Schema Enhancement}
\label{section:bot-enhancement}

\begin{figure*}[ht!]
    \includegraphics[width=0.9\linewidth]{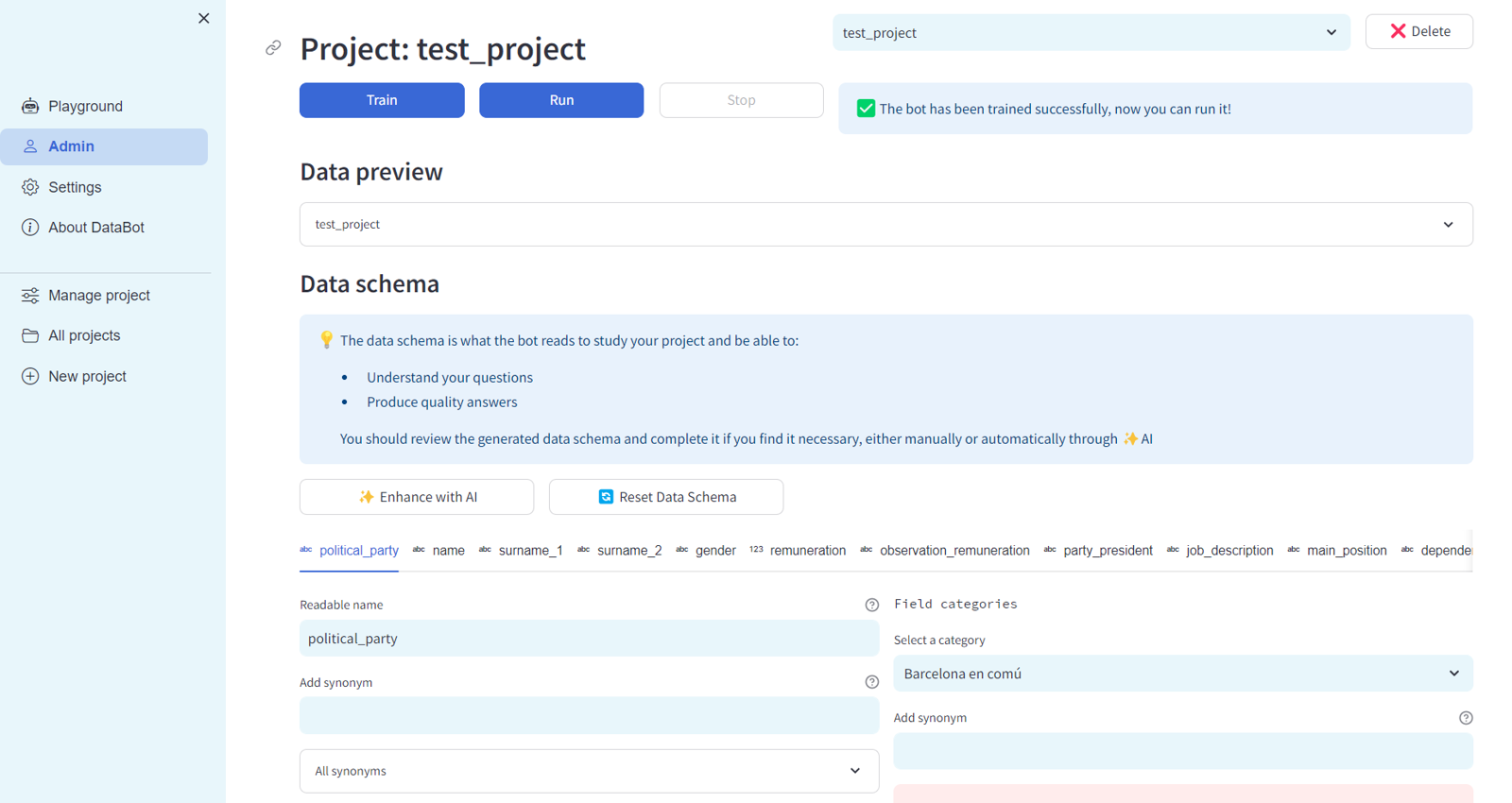}
    \caption{Screenshot of the admin User Interface, where bots can be executed and the data schemas can be enhanced.}
    \Description{Screenshot of the admin User Interface, where bots can be executed and the data schemas can be enhanced.}
    \label{fig:admin-ui}
\end{figure*}

Even though chatbots can be automatically created, they could  have limited success for some data sources. This could happen, for instance, if the user uses words very distant to those present in the data schema. Another common problem is the semantics of fields. Some fields may compose an address (street, number, city, etc.) or a date (year, month, day, hour). This is something users could ask about, but they are probably not aware of the internal structure of the dataset and could ask about fields that actually do not explicitly exist as such. 

Our bots' philosophy is based on being sure in the answers they provide. Therefore, increasing the bot comprehension or permission to understand what is unknown could derive in a much higher failure rate. This can be considered a limitation but it is also as a safety mechanism. However, we provide the user with the necessary tools to optionally enhance their bots' capabilities by enriching the automatically inferred data schema. As an example, the schema could be enriched by adding synonyms or creating new virtual columns (result of merging fields). These improvements can be done either manually by the bot creator or using a LLM to automatize the process, as shown in Figure \ref{fig:admin-ui}. The data schema model is designed to be easily extensible to add different knowledge components. Therefore, further work on this matter to augment the bot comprehension of the data can be done in the future.

\subsection{Bot Generation}
\label{section:bot-generation}

The bot generation phase takes the (potentially enriched) data schema and instantiates a set of predefined conversation patterns, gathered, improved and extended via several experiments with users, to generate the actual set of questions the bot will be trained on (\emph{i.e.}, the intents' training sentences). The training phase of the bot will vary depending on the chosen intent classifier component. The main idea behind this is that this component learns the kind of questions it will receive by seeing example (training) annotated sentences. In NLP, this problem is known as Text Classification.
On top of this core component, the generator will add the fallback mechanism and other auxiliary conversations and components to create a fully functional bot.

\subsubsection{Intents}

The generated bots will contain a set of predefined intents, whose training sentences will be generated from a template bundle and completed with the data schema information\footnote{\url{https://github.com/BESSER-PEARL/databot/blob/main/src/app/bot/library/intents.json}}. These intents have been designed to suit as many datasets as possible and to match any dataset query that involves the columns and their values as embedded parameters. These queries mainly (but not only) include any SQL `select' statement (in an equivalent natural language form), column comparisons, column or value frequencies, etc., regarding the tabular answer intents, and histogram, boxplot, bar, pie or line chart generation (among others) regarding the chart answer intents.

The advantage of this conversational model is that it is easily extensible with further intents allowing the integration of new bot capabilities by just defining the new intent and its proper response. 

\subsubsection{Entities}

The bots contain a set of entities used to recognize relevant elements within intents through their parameters. These entities consist of a set of values, each of them containing an optional set of synonyms. In our context, the parameters the bots must recognize are mainly elements relative to the data they serve, like field names or values. In other words, their content depends on the data content. There are other data-independent entities such as operators (\emph{e.g.}, `maximum' or `minimum') or row names (\emph{e.g.}, `row' or `entry', though the user can add domain-specific row names, like `person' or `employee')

\begin{figure*}[ht!]
    \includegraphics[width=0.9\linewidth]{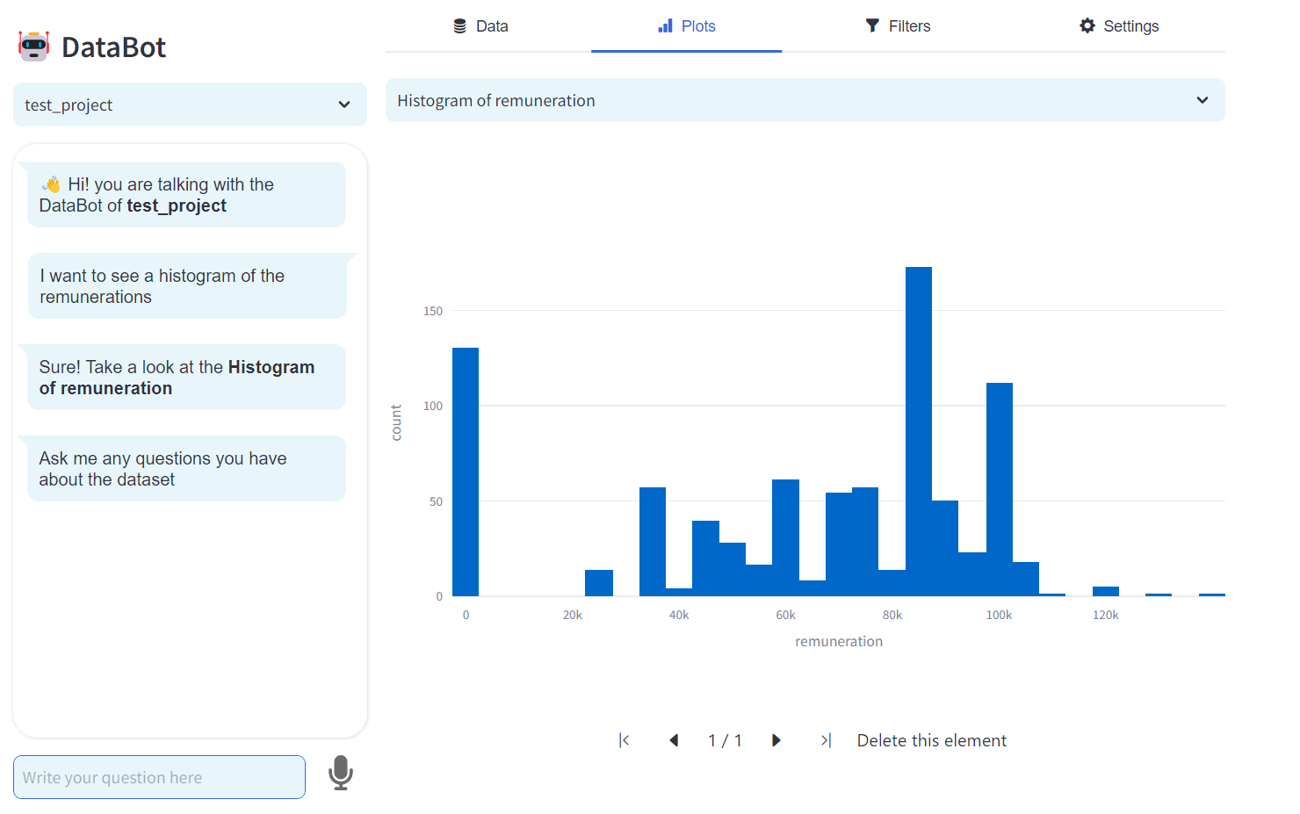}
    \caption{Screenshot of the interactive dashboard showing a graphic answer (a histogram) generated by the bot.}
    \Description{Screenshot of the interactive dashboard showing a graphic answer (a histogram) generated by the bot.}
    \label{fig:plot-answer}
\end{figure*}

\begin{figure*}[ht!]
    \includegraphics[width=0.9\linewidth]{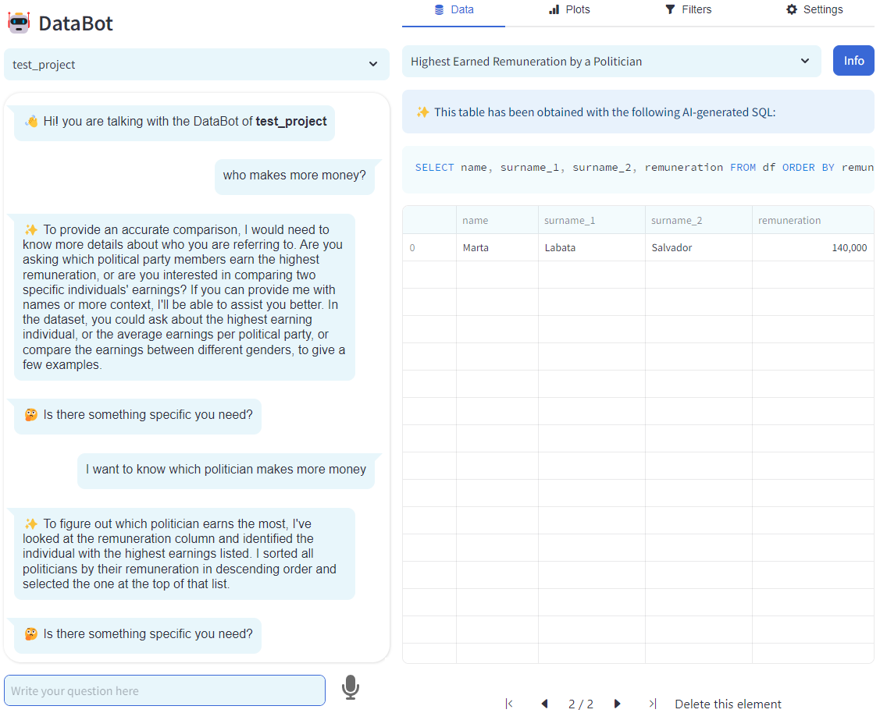}
    \caption{Screenshot of the interactive dashboard showing a tabular answer generated by the bot through the fallback mechanism powered by GPT-4. On the top-right side, there is an information box indicating that the  displayed answer has been obtained after running an AI-generated SQL statement, and the actual SQL is also shown.}
    \Description{Screenshot of the interactive dashboard showing a tabular answer generated by the bot through the fallback mechanism powered by GPT-4. On the top-right side, there is an information box indicating that the  displayed answer has been obtained after running an AI-generated SQL statement, and the actual SQL is also shown.}
    \label{fig:llm-answer}
\end{figure*}

\subsection{Using the Generated Bots}

Once admin users are ready to generate a chatbot, they can go ahead just by pressing the \emph{Train \& Run} buttons to (locally) deploy the bot. It will be available in the playground tab of the platform. The playground is the UI for the chatbot end-user (\emph{e.g.}, the citizen). It offers an interactive dashboard with a chat box on the left side of the canvas, together with a text input box and a voice input button. On the right side, there is the dashboard itself, composed of a set of tabs aimed at organizing the chatbot-generated content and configuration options. It is also possible to create filters, restricting the search space of the bot when generating an answer (\emph{e.g.}, filter by gender, before some date or with some numeric field lower than a threshold). Figures \ref{fig:plot-answer} and \ref{fig:llm-answer} show two different interactions with a chatbot that generated graphical and tabular answers, respectively.

\section{Tool Implementation}
\label{section:tool-support}

All the tool components are written in Python. The UI relies on Streamlit, a Python library for GUI development. The generated chatbots rely on the BESSER Bot Framework\footnote{\url{https://github.com/BESSER-PEARL/BESSER-Bot-Framework}}, a Python open-source bot development tool for its runtime execution. All the data management is performed with Pandas, the \emph{de facto} Python library for this type of task. Finally, for all the LLM-based tasks (data schema enhancement, intent classification and English-to-SQL fallback) we rely on OpenAI's GPT-4 through its API. All the source code of the DataBot platform is freely available on GitHub.

The design of the conversation patterns has been the result of three preliminary experiments with datasets from the city of Barcelona\footnote{\url{https://opendata-ajuntament.barcelona.cat/data/en/dataset/carrecs-electes-comissionats-i-gerents}}, the Catalan Government \footnote{\url{https://dadesobertes.seu-e.cat/dataset/ge-p-pressupostos-per-programes-detallat}} and the LIS Cross-National Data Center in Luxembourg\footnote{\url{https://www.lisdatacenter.org/wp-content/uploads/files/access-key-workbook.xlsx}}, where we deployed a chatbot for each dataset and gathered the user interactions (both from the data owners and the citizens) to refine the intents the bots should recognize.

\section{Related Work}
\label{section:related-work}

In this section, we compare our approach with other works focusing on the exploration and exploitation of tabular data by non-technical people.

A first group of works (\emph{e.g.}, Socrata\footnote{\url{https://dev.socrata.com/}}) focuses on the generation of charts and interactive dashboards \cite{wu2021multivision} to help users filter and view the data they want. 
However, while really useful to see trends and global data perspectives, these works cannot be used to answer concrete adhoc questions on specific aspects of the data.

Other approaches opt for a direct English-to-SQL translation when querying tabular data, such as \cite{TheNLPtoSQLWeUse, abraham2022tablequery, raffel2020exploring}. 
The major concern with these ``uncontrolled'' translations is that they can generate wrong queries and therefore come up with wrong factual answers. This latter issue is also the main concern with generative chatbots based on LLMs that could also be used to chat with the citizens. They tend to ``hallucinate'' and invent facts, which is something too risky for a public-facing chatbot, especially for a government administration \cite{shwartz2022tabular}.

More similar to our efforts, a couple of Proof of Concepts of intent-based chatbots used for open data have been published \cite{porreca2018accessing, CantadorVCB21}. Both bots were manually created. This is in contrast with our approach where bots are automatically generated. 
An exception is \cite{castaldo2019conversational} where the bot generation is semi-automated but it requires a mandatory and extensive annotation process while we focus more on a scalable approach able to generate chatbots with no human intervention if so desired. Generation of chatbots from other types of data sources like APIs \cite{HamzaAPI, VMAPIChatbot},  web pages \cite{chitto2020automaticwebchatbot}, knowledge graphs \cite{AvilaFMV20} or even software designs \cite{ChatbotsForModels} has also been explored and some of their ideas could be exploited as well for tabular data. However, we did not find any existing solution for automatic generation of intent-based chatbots from tabular datasets that we could compare with this work.

Another approach being widely used to query data with LLMs is Retrieval-Augmented Generation (RAG) \cite{NEURIPS2020_6b493230}. While being useful for unstructured data sources like plain text, RAG has some limitations to query tabular data. Some queries may need to retrieve an entire column (\emph{e.g.}, to sum all the values), and RAG's methodology consists of selecting relevant parts of the data to send them to a LLM as context so it can easily find the answer. This is not the best solution for tabular data since it is not scalable. Furthermore, LLMs still can suffer hallucinations.

To sum up, we believe our approach proposes a novel combination of strategies to mix the best of both worlds (intent-based and LLM-based chatbots) and opens the door to a more massive use of chatbots for tabular data thanks to our automatic generation strategy.

\section{Conclusions and Further Work}
\label{section:conclusions-and-future-work}

We have presented a new tool to automatically generate chatbots from tabular data sources to help non-technical users explore this type of data. This is especially useful in the current trend towards more transparency and openness in the public administration, with more and more open data sources released each day. Our chatbots encode a significant number of potential questions users may want to ask the data. Such questions are automatically generated based on an initial analysis of the structure and content of the data source. 

As further work, we plan to enrich the training of the chatbots with the use of ontologies. The idea would be to map the data schema to ontological concepts to be able to consider more semantic information in the training. We also plan to extend the set of conversation patterns including questions on the validity, origin and possible biases of the data \cite{DescribeML}.

\begin{acks}
This project is supported by the Luxembourg National Research Fund (FNR) PEARL program, grant agreement 16544475.
This work has been partially funded by the Spanish government (PID2020-114615RB-I00/AEI/10.13039/501100011033, project LOCOSS).
\end{acks}

\bibliographystyle{ACM-Reference-Format}
\bibliography{references}





\end{document}